\documentclass[11pt]{article}

\usepackage{jmlr2e}

\usepackage{graphics} % for pdf, bitmapped graphics files
\usepackage{epsfig} % for postscript graphics files
\usepackage{times} % assumes new font selection scheme installed
\usepackage{amsmath} % assumes amsmath package installed
\usepackage{amssymb}  % assumes amsmath package installed
\usepackage{psfrag}
\usepackage{graphicx}
\usepackage{float}
\newcommand{\comment}[1]{}

\providecommand{\definitionname}{Definition}
\providecommand{\theoremname}{Theorem}

\newcommand\mycomment[1]{} 			     % remove all comments and TODOs

\newcommand{\ignore}[1]{}
\ShortHeadings{Opening the black box of Deep Neural Networks via Information}{Schwartz-Ziv and Tishby}
\firstpageno{1}

\begin{document}

\title{Opening the black box of Deep Neural Networks\\ via Information}

\author{\name Ravid Schwartz-Ziv \email ravid.ziv@mail.huji.ac.il \\
       \addr Edmond and Lilly Safra Center for Brain Sciences\\
       The Hebrew University of Jerusalem\\
       Jerusalem,  91904, Israel
%       \AND 
%       \name Noga Zaslavsky \email noga.zaslavsky@mail.huji.ac.il \\
%       \addr Edmond and Lilly Safra Center for Brain Sciences\\
%       The Hebrew University of Jerusalem\\
%       Jerusalem,  91904, Israel
       \AND
       \name Naftali Tishby\footnote{Corresponding author} \email tishby@cs.huji.ac.il \\
       \addr School of Engineering and Computer Science\\ and 
       Edmond and Lilly Safra Center for Brain Sciences\\
       The Hebrew University of Jerusalem\\
       Jerusalem,  91904, Israel
       }

\editor{ICRI-CI}

\maketitle

\begin{abstract}%   <- trailing '%' for backward compatibility of .sty file
Despite their great success, there is still no comprehensive theoretical understanding of learning with Deep
Neural Networks (DNNs) or their inner organization.
Previous work [\citet{DBLP:journals/corr/TishbyZ15}] proposed to analyze DNNs in the \textit{Information
Plane}; i.e., the plane of the Mutual Information values that each layer preserves on the input and output variables. They suggested that the goal of the network is to optimize the Information Bottleneck (IB) tradeoff between compression and prediction, successively, for each layer.

In this work we follow up on this idea and demonstrate the effectiveness of the Information-Plane visualization of DNNs. 
Our main results are: (i) most of the training epochs in standard DL are spent on {\emph compression} of the input to efficient representation and not on fitting the training labels. (ii) The representation compression phase begins when the training errors becomes small and the Stochastic Gradient Decent (SGD) epochs change from a fast drift to smaller training error into a stochastic relaxation, or random diffusion, constrained by the training error value. (iii)  The converged layers lie on or very close to the Information Bottleneck (IB) theoretical bound, and the maps from the input to any hidden layer and from this hidden layer to the output satisfy the IB self-consistent equations. This generalization through noise mechanism is unique to Deep Neural Networks and absent in one layer networks. (iv) The training time is dramatically reduced when adding more hidden layers. Thus the main advantage of the hidden layers is computational. This can be explained by the reduced relaxation time, as this it scales super-linearly (exponentially for simple diffusion) with the information compression from the previous layer. (v) As we expect critical slowing down of the stochastic relaxation near phase transitions on the IB curve, we expect the hidden layers to converge to such critical points.\footnote{This paper was done with the support of the Intel Collaborative Research institute for Computational Intelligence (ICRI-CI) and is part of the “Why \& When Deep Learning works: looking inside Deep Learning” ICRI-CI paper bundle.} 
\end{abstract}


\newpage

\section{Introduction}
\label{Introduction}

In the last decade, deep learning algorithms have made remarkable progress
on numerous machine learning tasks and dramatically improved the state-of-the-art in many practical areas
 [\citet{DBLP:journals/corr/abs-1303-5778,DBLP:journals/corr/ZhangL15,DBLP:journals/corr/abs-1207-0580,DBLP:journals/corr/HeZRS15,natureDeepLeraning}].

Despite their great success, there is still no comprehensive understanding of the optimization process 
or the internal organization of DNNs, and they are often criticized for being used as mysterious "black boxes" 
[e.g., \citet{probes2016}].

\begin{figure}[h]
\begin{centering}
\includegraphics[scale = 0.3]{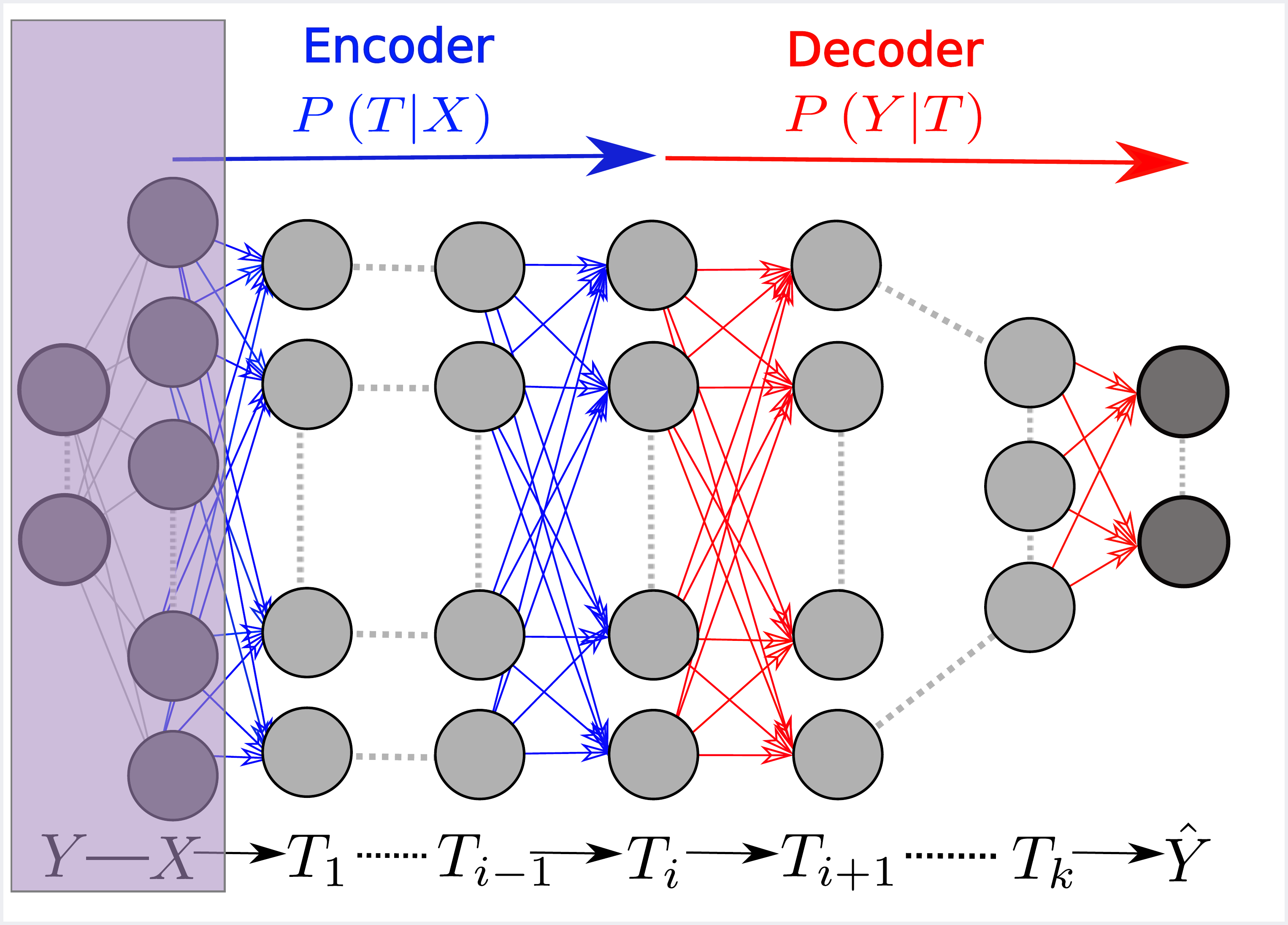}
\par\end{centering}
\caption{The DNN layers form a Markov chain of successive internal representations of the input layer $X$. Any representation of the input, $T$, is defined through an encoder, $P(T|X)$, and a decoder $P(\hat{Y}|T)$, and can be quantified by its \emph{information plane} coordinates: $I_X= I(X;T)$ and $I_Y=I(T;Y)$. The Information Bottleneck bound characterizes the optimal representations, which maximally compress the input $X$, for a given mutual information on the desired output $Y$.   After training, the network receives an input $X$, and successively processes it through the layers, which form a Markov chain, to the predicted output $\hat{Y}$. $I(Y;\hat{Y})/I(X;Y)$ quantifies how much of the relevant information is captured by the network.
}
\label{DNN-layers}
\end{figure}

In \citet{DBLP:journals/corr/TishbyZ15}, the authors noted that layered neural networks form a Markov chain of successive representations of the input layer and suggested studying them in the \textit{Information Plane} 
 - the plane of the Mutual Information values of any other variable with the input variable $X$ and desired output variable $Y$ (Figure \ref{DNN-layers}).  The rationale for this analysis was based on the invariance of the mutual information to invertible re-parameterization and on the Data Processing Inequalities along the Markov chain of the layers. Moreover, they suggested that optimized DNNs layers should approach the Information Bottleneck (IB) bound [\citet{DBLP:journals/corr/Tishby1999}] of the optimal achievable representations of the input $X$. 

In this paper we extend their work and demonstrate the effectiveness of the 
visualization of DNNs in the information plane for a better understating of the training dynamics, 
learning processes, and internal representations in Deep Learning (DL). 

Our analysis reveals, for the first time to our knowledge, that the Stochastic Gradient Decent (SGD) optimization, commonly used  in Deep Learning, has two different and distinct phases: empirical error minimization (ERM) and representation compression. 
These phases are characterized by very different signal to noise ratios of the stochastic gradients in every layer. In the ERM phase the gradient norms are much larger than their stochastic fluctuations, resulting in a rapid increase in the mutual information on the label variable $Y$. In the compression phase, the fluctuations of the gradients are much larger than their means,  and the weights change essentially as Weiner processes, or random diffusion, with a very small influence of the error gradients. 

This phase is marked by a slow representation compression, or reduction of the mutual information on the input variable $X$.   
In our experiments, most of the optimization epochs are spent on compressing the internal representations under the training error constraint. This compression occurs by the SGD without any other explicit regularization or sparsity, and - we believe - is largely responsible for the absence of overfitting in DL. This observation also suggests that there are many (exponential in the number of weights) different randomized networks with essentially optimal performance. Hence the interpretation of a single neuron (or weight) in the layers is practically meaningless. 

We then show that the optimized layers, for large enough training samples, lie on or very close to the optimal IB bound, resulting in a self-consistent relationship between the encoder and decoder distributions for each layer (Figure \ref{DNN-layers}). The optimized hidden layers converge along special lines in the information plane, and move up in the plane as we increase the training sample size.  
Finally, the diffusive nature of the SGD dynamics provides a new explanation for the computational benefit of the hidden layers. 

\subsection{Summary of results and structure of the paper}
Our analysis gives the following new results: (i) the Stochastic Gradient Decent (SGD) optimization
has two main phases. In the first and shorter phase the layers increase the information on the labels (fitting), 
while in the second and much longer phase the layer reduce the information on the input (compression phase). 
We argue that the second phase amounts to a stochastic relaxation (diffusion) that maximizes the conditional entropy of the layers subject to the empirical error constraint.  (ii) The converged layers lie on or very close to the IB theoretical bound, for different values of the tradeoff parameter, and the maps from the input to to each layer (encoder) and from the layer to the output (decoder) satisfy the IB self-consistent optimality conditions. (iii) The main advantage of the hidden layers is computational, as they dramatically reduce the stochastic relaxation times. (iv) The hidden layers appear to lie close to critical points on the IB bound, which can be explained by critical slowing down of the stochastic relaxation process. 

In section \ref{Sec:DL-IT} we describe the Information Theoretic aspects of Deep Learning, the relevant properties of mutual information, the Information Bottleneck framework, and the visualization of DNNs in the Information Plane. The main part of the paper is in section \ref{Sec:Experiments} where we describe our experimental setting a list of new insights they provide about the dynamics of the SGD optimization and the mechanism in which this stochastic dynamics achieves the IB bound. We also discuss the benefit of adding hidden layers and the special locations of the final layers in the information plane. We conclude with a discussion of further critical issues in section \ref{Sec:discussion}.  
\section{Information Theory of Deep Learning}
\label{Sec:DL-IT}
In supervised learning we are interested in good representations, $T(X)$, of the input patterns  $x\in X$, that enable good predictions of the label  $y \in Y$. Moreover, we want to efficiently learn such representations from an empirical sample of the (unknown) joint distribution $P(X,Y)$, in a way that provides good generalization.

DNNs and Deep Learning generate a Markov chain of such representations, the hidden layers, by minimization of the empirical error over the weights of the network, layer by layer. This optimization takes place via stochastic gradient descent (SGD), using a noisy estimate of the gradient of the empirical error at each weight, through back-propagation.  

Our first important insight is to treat the whole layer, $T$, as a single random variable, characterized by its encoder, $P(T|X)$, and decoder, $P(Y|T)$ distributions. As we are only interested in the information that flows through the network, invertible transformations of the representations, that preserve information, generate equivalent representations even if the individual neurons encode entirely different features of the input. For this reason we quantify the representations by two numbers, or order parameters, that are invariant to any invertible re-parameterization of $T$, the mutual information of $T$ with the input $X$ and the desired output $Y$.

\ignore{ We argue here that SGD works in very different ways during the two distinct phases of the optimization process, both are critical for the success of Deep Learning. During the first phase, which lasts in our experiments a few hundred epochs, the noisy gradient has a narrow distribution around the true training error gradient, and the error decreases rapidly. In the second phase, for thousands additional epochs, the gradient distribution width is much larger than it's mean and the SGD induces random perturbations which maximizes the entropy of the weights distribution, for every hidden layer. 
This random walk, or stochastic relaxation process, \ravidcomment{ I think we need to explain these terms somewhere} converge to a different (special) distribution for each hidden layer, and each layer forms a different efficient representation of the input variable.}
 
Next, we quantify the quality of the layers by comparing them to the information theoretic optimal representations, the Information Bottleneck representations, and then describe how Deep Learning SGD can achieve these optimal representations. 

\subsection{Mutual Information}
Given any two random variables, $X$ and $Y$, with a joint distribution $p(x,y)$, their Mutual Information is defined as:
%\small{
\begin{align}
I(X;Y) & = D_{KL}[p(x,y)||p(x)p(y)] = \sum_{x\in X, y\in Y} p(x,y) \log\left(\frac{p\left(x,y\right)}{p\left(x\right)p\left(y\right)}\right)\\
 & = \sum_{x\in X, y\in Y}p\left(x,y\right)\log\left(\frac{p\left(x|y\right)}{p\left(x\right)}\right)
 = H(X)-H(X|Y) ~,
 \label{MI}
\end{align}
%}
\normalsize
where $D_{KL}[p||q]$ is the Kullback-Liebler divergence of the distributions $p$ and $q$, and $H(X)$ and $H(X|Y)$ are the entropy and conditional entropy of $X$ and $Y$, respectively. 

The mutual information quantifies the number of relevant bits that the input variable $X$ contains about the label $Y$, on average. The optimal learning problem can be cast as the construction of an \textit{optimal encoder} of that relevant information via an efficient representation - a minimal sufficient statistic of $X$ with respect to $Y$ - if such can be found. A minimal sufficient statistic can enable the \textit{decoding} of the relevant information with the smallest number of  binary questions (on average);  i.e., an optimal code. The connection between mutual information and minimal sufficient statistics is discussed in \ref{sec:IB}.

Two properties of the mutual information are very important in the context of DNNs. 
The first is its invariance to invertible transformations:
\begin{equation}
\label{eqn:invariance}
I\left(X;Y\right)=I\left(\psi(X);\phi(Y))\right)
\end{equation}
for any invertible functions $\phi$ and $\psi$.
 
The second is the Data Processing Inequality (DPI) [\citet{Cover:2006}]: 
for any 3 variables that form a Markov chain $X\rightarrow Y \rightarrow Z$,
\begin{equation}
I\left(X;Y\right) \ge I(X;Z) ~.
\end{equation}
 
%\subsection{Learning efficient representations}

\subsection{The Information Plane} 
\label{I-Plane}
%\normalsize 
Any representation variable, $T$, defined as a (possibly stochastic) map of the input $X$, is characterized by its joint distributions with $X$ and $Y$, or by its encoder and decoder distributions, $P(T|X)$ and  $P(Y|T)$, respectively. 
Given $P(X;Y)$, $T$ is uniquely mapped to a point in the information-plane with coordinates 
$\left(I(X;T),I(T;Y)\right)$. When applied to the Markov chain of a K-layers DNN, with $T_i$ denoting the $i^{th}$ hidden layer as a single multivariate variable (Figure \ref{DNN-layers}), the layers are mapped to $K$ monotonic connected points in the plane - henceforth  a unique \textit{information path} -  which satisfies the following DPI chains:
%\vskip -0.1in
%\small
\begin{align}
I(X;Y) & \ge I(T_1;Y) \ge I(T_2;Y) \ge ...\ge I(T_k;Y) \ge I(\hat{Y};Y)\\
H(X) & \ge I(X;T_1) \ge I(X;T_2) \ge ... \ge I(X;T_k) \ge I(X;\hat{Y}) .
 \label{DPI-1}
\end{align}
%\normalsize
Since layers related by invertible re-parametrization appear in the same point, each information path in the plane corresponds to many different DNN's, with possibly very different architectures.

\subsection{The Information Bottleneck optimal representations}
\label{sec:IB}

What characterizes the optimal representations of $X$ w.r.t. $Y$? 
The classical notion of minimal sufficient statistics provide good candidates for optimal representations. Sufficient statistics, in our context, are maps or partitions of $X$, $S(X)$, that capture all the information that $X$ has on $Y$. Namely, $I(S(X);Y)=I(X;Y)$ [\citet{Cover:2006}]. 

Minimal sufficient statistics, $T(X)$, are the simplest sufficient statistics and induce the coarsest sufficient partition on $X$. In other words, they are functions of any other sufficient statistic. A simple way of formulating this is through the Markov chain: $Y\rightarrow X \rightarrow S(X) \rightarrow T(X)$, which should hold for a minimal sufficient statistics $T(X)$ with any other sufficient statistics $S(X)$. Using the DPI, we can cast it into a constrained optimization problem:
\begin{equation}
T(X) = \arg \min_{S(X): I(S(X);Y)=I(X;Y)} I(S(X);X) .
\end{equation}

Since exact minimal sufficient statistics only exist for very special distributions, (i.e., exponential families), \citet{DBLP:journals/corr/Tishby1999} relaxed this optimization problem by first allowing the map to be stochastic, defined as an encoder $P(T|X)$, and then, by allowing the map to capture \emph{as much as possible} of $I(X;Y)$, not necessarily all of it.

This leads to the \textit{Information Bottleneck} (IB) tradeoff [\citet{DBLP:journals/corr/Tishby1999}], which 
provides a computational framework for finding approximate minimal sufficient statistics, or the optimal tradeoff between compression of $X$ and prediction of $Y$.  Efficient representations are approximate minimal sufficient statistics in that sense.

If we define $t\in T$ as the compressed representations of $x\in X$, the
representation of $x$ is now defined by the mapping $p\left(t|x\right)$.
This Information Bottleneck tradeoff is formulated by the following optimization problem, carried independently for the distributions, $p(t|x), p(t), p(y|t)$, with the Markov chain: $Y\rightarrow X \rightarrow T$,
\begin{equation}
\min_{p\left(t|x\right),p\left(y|t\right),p\left(t\right)}\left\{ I\left(X;T\right)-\beta I\left(T;Y\right)\right\} ~.
\end{equation}

The Lagrange multiplier $\beta$ determines the level of relevant information captured by the representation $T$, $I(T;Y)$, which is directly related to the error in the label prediction from this representation.  The (implicit) solution to this problem is given by three IB self-consistent equations:
\begin{equation}
\label{eqn:IB}
\begin{cases}
p\left(t|x\right)=\frac{p\left(t\right)}{Z\left(x;\beta\right)}\exp\left(-\beta D_{KL}\left[p\left(y|x\right)\parallel p\left(y|t\right)\right]\right)\\
p\left(t\right)=\sum_{x}p\left(t|x\right)p\left(x\right)\\
p\left(y|t\right)=\sum_{x}p\left(y|x\right)p\left(x|t\right)~,
\end{cases}
\end{equation}
where $Z\left(x;\beta\right)$ is the normalization function. 
These equations are satisfied along the \emph{information curve}, which is a monotonic concave line of optimal representations that separates the achievable and unachievable regions in the information-plane. For \emph{smooth} $P(X,Y)$ distributions; i.e., when $Y$ is not a completely deterministic function of $X$, 
%\mycomment{do we need to explain that?  it can be tricky},
the information curve is strictly concave with a unique slope, $\beta^{-1}$, at every point, and a finite slope at the origin. 
In these cases $\beta$ determines a single point,  
%$\left(I_{X}^{\beta}=I_{\beta}(T;X), I_{Y}^{\beta}=I_{\beta}(T;Y)\right)$ 
%\ravidcomment{It's hard to read it like that, maybe without the brakets?  }
on the information curve with specified \emph{encoder},$P^{\beta}(T|X)$, and \emph{decoder},  $P^{\beta}(Y|T)$, distributions that are related through Eq.(\ref{eqn:IB}).  
For deterministic networks, we consider the sigmoidal output of the neurons as probabilities, consistent with the commonly used cross-entropy or log-loss error in the  stochastic optimization.  
The rest of our analysis is restricted to these distributions and networks.

\subsection{The crucial role of noise}
The invariance of the information measures to invertible transformations comes with a high cost. 
For deterministic functions, $y=f(x)$, the mutual information is insensitive to the complexity of the function $f(x)$ or the class of functions it comes from. This can be easily seen for finite cardinality $|X|$, as the mutual information is invariant to any random permutation of the patterns in $X$. If we have no information on the structure or topology of $X$, then even for a binary $Y$ there is no way to distinguish low complexity classes (e.g. functions with a low VC dimension) from highly complex classes (e.g. essentially random function with high mixing complexity, see: \citet{MoshkovichTishby17}), by the mutual information alone. This looks like bad news for understanding machine learning complexity using only information measures. 

There is, however, a fundamental cure to this problem. We can turn the function into a stochastic rule by adding (small) noise to the patterns.  Consider the stochastic version of the rule given by the conditional probabilities $p(y|x)$, with values not only 0 or 1 that are sensitive to the distance to the decision boundary in $X$. Moreover, we want this probability to the complexity of the decision boundary in the standard learning complexity sense. The simple example to such rules is the Perceptron, or single formal neuron, with the standard sigmoid output $p(y=1)=\psi(w\cdot x)$, with $\psi(x)=\frac{1}{1+\exp(-x)}$. In this case we can interpret the output of the sigmoid as the probability of the label $y=1$ and the function is a simple linear hyper-plane in the input space $X$ with noise determined by the width of the sigmoid function. The joint distribution $p_w(x,y)=\frac{p(x)}{1+\exp(y-{\bf w}\cdot {\bf x}+{b})}$, and the distribution of $p(y|x_i)$ for the training data $x_i$ spread in the simplex $[0,1]$ in a very informative way. 
The sufficient statistics in this case is the dot product ${\bf w}\cdot {\bf x}$ with precision (in bits)  determined by the dimensionality of ${\bf w}$ and the margin, or by the level of noise in the sigmoid function. With more general functions, $y=f_w(x)$, 
$p_w(x,y)=\frac{p(x)}{1+\exp(y-f_w(x))}$ and its learning complexity is determined by the complexity (VC or related complexity measures) of the function class $f_w(x)$.

The learning complexity is related to the number of relevant bits required from the input patterns $X$ for a good enough prediction of the output label $Y$, or the minimal $I(X;\hat{X})$ under a constraint on $I(\hat{X};Y)$ given by the IB. Without the stochastic spread of the sigmoid output this mutual information is simply the entropy $H(Y)$ independent of $f_w(x)$, and there is nothing in the structure of the points $p(y|x)$ on the simplex to hint to the geometry or learning complexity of the rule.

\subsection{Visualizing DNNs in the Information Plane}

As proposed by \citet{DBLP:journals/corr/TishbyZ15}, we study the \emph{information paths} of DNNs in the \emph{information plane}. This can be done when the underlying distribution, $P(X,Y)$, is known and the encoder and decoder distributions $P(T|X)$ and $P(Y|T)$ can be calculated directly. For "large real world" problems these distributions and mutual information values should be estimated from samples or by using other modeling assumptions. Doing this is beyond the scope of this work, but we are convinced that our analysis and observations are general, and expect the dynamics phase transitions to become even sharper for larger networks, as they are inherently based on statistical ensemble properties. Good overviews on methods for mutual information estimation can be found in \citet{Paninski:2003:EEM:795523.795524} and \citet {PhysRevE.69.066138}.

\ignore{Due to the invariance property of the mutual information, Eq. (\ref{eqn:invariance}), each point in the \textit{information plane} represents an ensemble of possible networks/layers with the same information efficiency but possibly very different connections.}
Our two order parameters, $I(T;X)$ and $I(T;Y)$, allow us to visualize and compare different network architectures in terms of their efficiency in preserving the relevant information in $P(X;Y)$.

By visualizing the paths of different networks in the \textit{information plane} we explore the following fundamental issues: 
\begin{enumerate}
\item The SGD layer dynamics in the \textit{Information plane}.
\item The effect of the training sample size on the layers.
\item What is the benefit of the hidden layers?
\item What is the final location of the hidden layers?
\item Do the hidden layers form optimal IB representations? 
%6. The optimal DNN architecture for a given task. 
\end{enumerate}

\section{Numerical Experiments and Results}
\label{Sec:Experiments}
\subsection{Experimental Setup }
For the numerical studies in this paper we explored fully connected feed-forward neural networks, with no other architecture constraints. 
We used standard DNN settings. 
The activation function of all the neurons was the hyperbolic tangent function, shifted to a sigmoidal function in the final layer. 
The networks were trained using SGD and the cross-entropy loss function, with no other explicit regularization. 
Unless otherwise noted, the DNNs used had up to 7 fully connected hidden layers, with widths:
  12-10-7-5-4-3-2 neurons (see Figure \ref{fig:gradients}). 
In our results below, layer 1 is the hidden layer closest to the input and the highest is the output layer.

To simplify our analysis, the tasks were chosen as binary decision rules which are invariant under $O(3)$ rotations of the sphere, with 12 binary inputs that represent 12 uniformly distributed points on a 2D sphere.
We tested other - non-symmetric - rules, but they had no effect on the results and conclusions of this paper (see supplementary material) . 
With such rules, the 4096 different patterns of the input variable $X$ are divided into 64 disjoint orbits of the rotation group. These orbits form a minimal sufficient partition/statistics for spherically symmetric rules [\citet{Kazhdan2003}].
  
\ignore{
Given a function $f$ defined on the sphere, we can obtain a rotation and reflection invariant representation  $\psi\left(f\right)$ by computing the spherical harmonic decomposition of the function 
\[
\textbf{$f\left(\theta,\phi\right)=\sum_{l\geq{0}}\sum_{m=-l}^{l}{a_{l}^{m}Y_{l}^{m}\left(\theta, \phi\right)}$}
\]
and calculate the energies ($L_2$ norms) of the frequency components.
 \[
\textbf{$\psi\left(f\right)=\left\lbrace\|a_{0}^{0}\|, \sqrt[â¢]{\|a_{1}^{-1}\|^2+\|a_{1}^{0}\|^2+\|a_{1}^{1}\|^2},...\right\rbrace$}
\]
We would refer the readers to \cite{Kazhdan2003} for good exposition on the matter.
}
To generate the input-output distribution, $P(X,Y)$, we calculated a spherically symmetric real valued function of the pattern
$f(x)$ (evaluated through its spherical harmonics power spectrum [\citet{Kazhdan2003}]) and compared it to a threshold, $\theta$, and apply a step $\Theta$ function to obtain a  $\{0,1\}$ label:  $y(x)=\Theta(f(x)-\theta)$. We then soften it to a stochastic rule through a standard sigmoidal function, 
$\psi(u)=1/(1+\exp(-\gamma u))$, as:
%\vskip -0.1in
\begin{equation}
\label{eq:rule}
p(y=1|x)=\psi(f(x)-\theta)~.
\end{equation}
%\vskip -0.1in
The threshold $\theta$ was selected such that $p(y=1)=\sum_x p(y=1|x)p(x) \approx 0.5$, with uniform $p(x)$. The sigmoidal gain $\gamma$ was high enough to keep the mutual information $I(X;Y)\approx0.99$ bits. 
%\mycomment{Ravid, please verify this!, This is true!}

\subsection{Estimating the Mutual Information of the Layers}

As mentioned above, we look at each of the layers $1\le i\le K$ in the network as a single variable $T_{i}$, and calculate the mutual information between each layer with the input and with the labels. 

In order to calculate the mutual Information of the network layers with the input and output variables, we binned 
the neuron's $arctan$ output activations into 30 equal intervals between -1 and 1. 
We then used these discretized values for each neuron in the layer, $t\in T_i$, to directly calculate the 
joint distributions, over the 4096 equally likely input patterns $x\in X$, $P(T_i,X)$ and 
$P(T_i,Y)=\sum_x P(x,Y)P(T_i|x)$, using the Markov chain $Y\rightarrow X \rightarrow T_i$ for every hidden layer.  
Using these discrete joint distributions we calculated the decoder and encoder mutual information, $I(X;T_i)$ and $I(T_i;Y)$, for each hidden layer in the network.  

We repeated these calculations with 50 different randomized initialization of the network's weights and different random selections of the training samples, randomly distributed according to the rule $P(X,Y)$ in Eq.(\ref{eq:rule}).

\ignore{
Let be $x\in X$ and $y\in Y$ two random variable with a joint distribution
$p\left(x,y\right)$, and a DNN with $N$ layers. As previously explained,
we can look on each of the layers $1\le i\le N$ in the network as
random variable $T_{i}$. Therefore, the mutual information between
the output of each layer and $X$ and the output of each layer and
$Y$ - $I\left(X;T_{i}\right)$ and $I\left(T_{i};Y\right)$ respectively
can be calculated by - 
\begin{align}
I_{Y}^{i}=I(Y;T_{i}) & =\sum_{y\in Y,t\in T_{i}}p\left(y,t\right)\log\left(\frac{p\left(y,t\right)}{p\left(y\right)p\left(t\right)}\right)\\
 & =\sum_{y\in Y,t\in T_{i}}p\left(y|t\right)p\left(t\right)\log\left(\frac{p\left(y|t\right)}{p\left(y\right)}\right)
\end{align}

and 

\begin{align}
I_{X}^{i}=I(X;T_{i}) & =\sum_{x\in X,t\in T_{i}}p\left(x,t\right)\log\left(\frac{p\left(x,t\right)}{p\left(x\right)p\left(t\right)}\right)\\
 & =\sum_{x\in X,t\in T_{i}}p\left(x|t\right)p\left(t\right)\log\left(\frac{p\left(x|t\right)}{p\left(x\right)}\right)
\end{align}
}

\subsection{The dynamics of the training by Stochastic-Gradient-Decent}

\begin{figure}[ht]
%\vskip 0.05in
\centerline{\includegraphics[width=\textwidth]{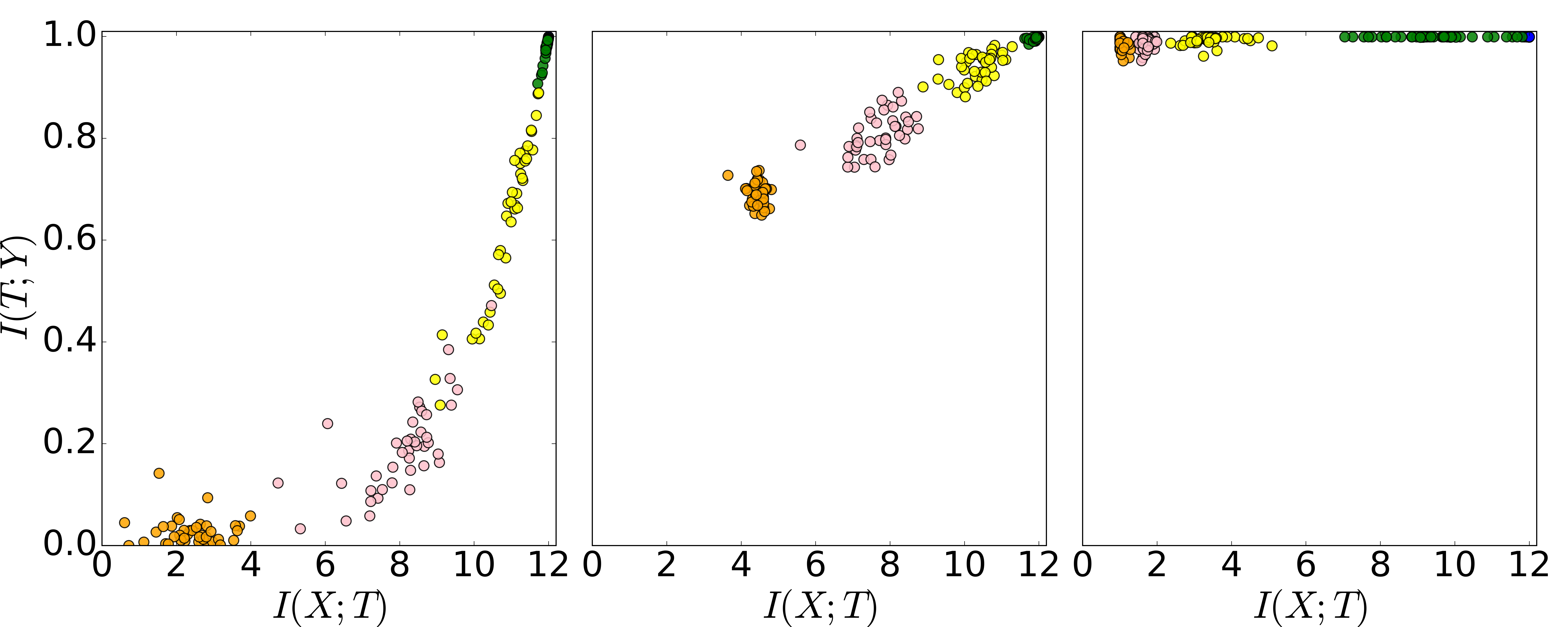}}
\caption{Snapshots of layers (different colors) of 50 randomized networks during the SGD optimization process in the \textit{information plane} (in bits):  \textbf{left} - with the initial weights;  \textbf{center} -  at 400 epochs; \textbf{right} -  after 9000 epochs.   The reader is encouraged to view the full videos of this optimization process in the \textit{information plane} at \textit{https://goo.gl/rygyIT} and \textit{https://goo.gl/DQWuDD}.}
\label{opt_process}
\vskip 0.1in
\end{figure}

To understand the dynamics of the network SGD optimization, we plot 
$I_X=I(X;T_i)$ and $I_Y= I(T_i;Y)$  for each layer for 50 different randomized initializations, with different randomized training samples. 
Figure \ref{opt_process} depicts the layers (in different colors) of all the 50 networks, trained with a randomized 85\% of the input patterns, in the information plane . 

\subsection{The two optimization phases in the Information Plane}

\begin{figure}[th]
\begin{centering}
\includegraphics[width=1.05\textwidth, height=5.8cm, scale = 0.6]{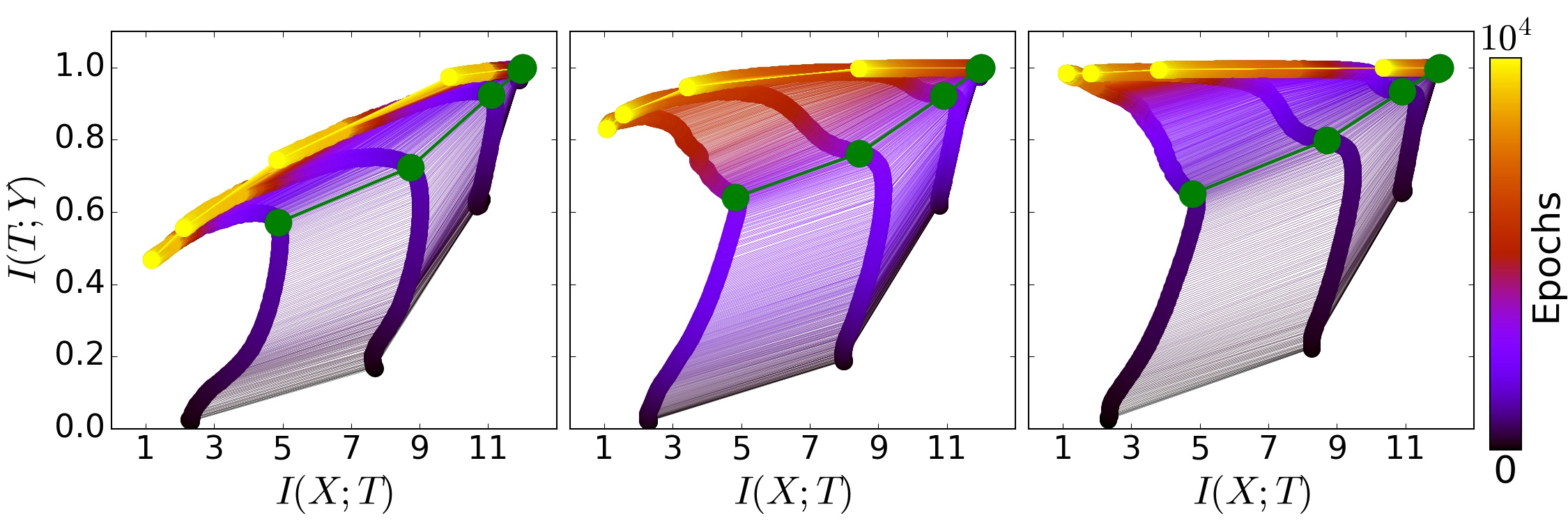}
\par\end{centering}
\caption{The evolution of the layers with the training epochs in the information plane, for different training samples. On the left - 5\% of the data, middle - 45\% of the data, and right - 85\% of the data. The colors indicate the number of training epochs with Stochastic Gradient Descent from 0 to 10000. The network architecture was fully connected layers, with widths: input=12-10-8-6-4-2-1=output. The examples were generated by the spherical symmetric rule described in the text. The green paths correspond to the SGD drift-diffusion phase transition - grey line on Figure \ref{fig:gradients} 
%[ToDo: INCREASE FONTS, ADD GREEN PATHS].
}
\label{network_epochs}
\vskip -0.1in
\end{figure}

As can be seen, at the beginning of the optimization the deeper layers of the randomly-initialize network fail to preserve the relevant information, and there is a sharp decrease in $I_Y$ along the path.  During the SGD optimization the layers first increase $I_Y$,  and later significantly decrease $I_{X}$, thus compressing the representation. 
Another striking observation is that the layers of the different randomized networks seem to follow very similar paths during the optimization and eventually converge to nearby points in the \textit{information plane}. Hence it is justified to average over the randomized networks, and plot the average layer trajectories in the plane, as shown in Figure \ref{network_epochs}.

On the right are the average network layers trajectories, when trained on random labeled samples of 85\% of the patterns, and on the left the same trajectories when under-trained on samples of only 5\% of the patterns. The middle depicts an intermediate stage with samples of 45\% of the data. Note that the mutual information is calculated with the full rule distribution, thus $I(T;Y)$ corresponds to the test, or generalization, error.  
The two optimization phases are clearly visible in all cases. During the fast - ERM - phase, which takes a few hundred  epochs, the layers increase the information on the labels (increase $I_{Y}$) while preserving the DPI order (lower layers have higher information). In the second and much longer training phase the layers' information on the input, $I_X$, decreases and the layers lose irrelevant information until convergence (the yellow points). 
We call this phase the \emph{representation compression} phase. 

While the increase of $I_{Y}$ in the ERM phase is expected from the cross-entropy loss minimization, the surprising compression phase requires an explanation.  There was no explicit regularization that could simplify the representations, such as $L1$ regularization, and there was no sparsification or norm reduction of the weights (see appendix). 
%\ravidcomment{Add this figure}). 
We observed the same two-phase layer trajectories in other problems, without symmetry or any other special structure. 
Thus it seems to be a general property of SGD training of DNNs, but it should be verified on larger problems. The observation and explanation of this phase is our main result.

Whereas the ERM phase looks very similar for both small (5\%) and large (85\%) training sample sizes, the compression phase significantly reduced the layers' label information in the small sample case, but with large samples the label information mostly increased. This looks very much like overfitting the small sample noise, which can be avoided with early stopping methods [\citet{Larochelle:2009:EST:1577069.1577070}]. Note, however, that this overfitting is largely due to the compression phase, which simplifies the layers' representations but also loses relevant information.  Understanding what determines the convergence points of the layers in the \textit{information plane}, for different training data sizes, is an interesting theoretical goal we currently investigate.

\subsection{The drift and diffusion phases of SGD optimization}

\begin{figure}[ht]
\vskip -0.0in
\centerline{\includegraphics[width=\textwidth]{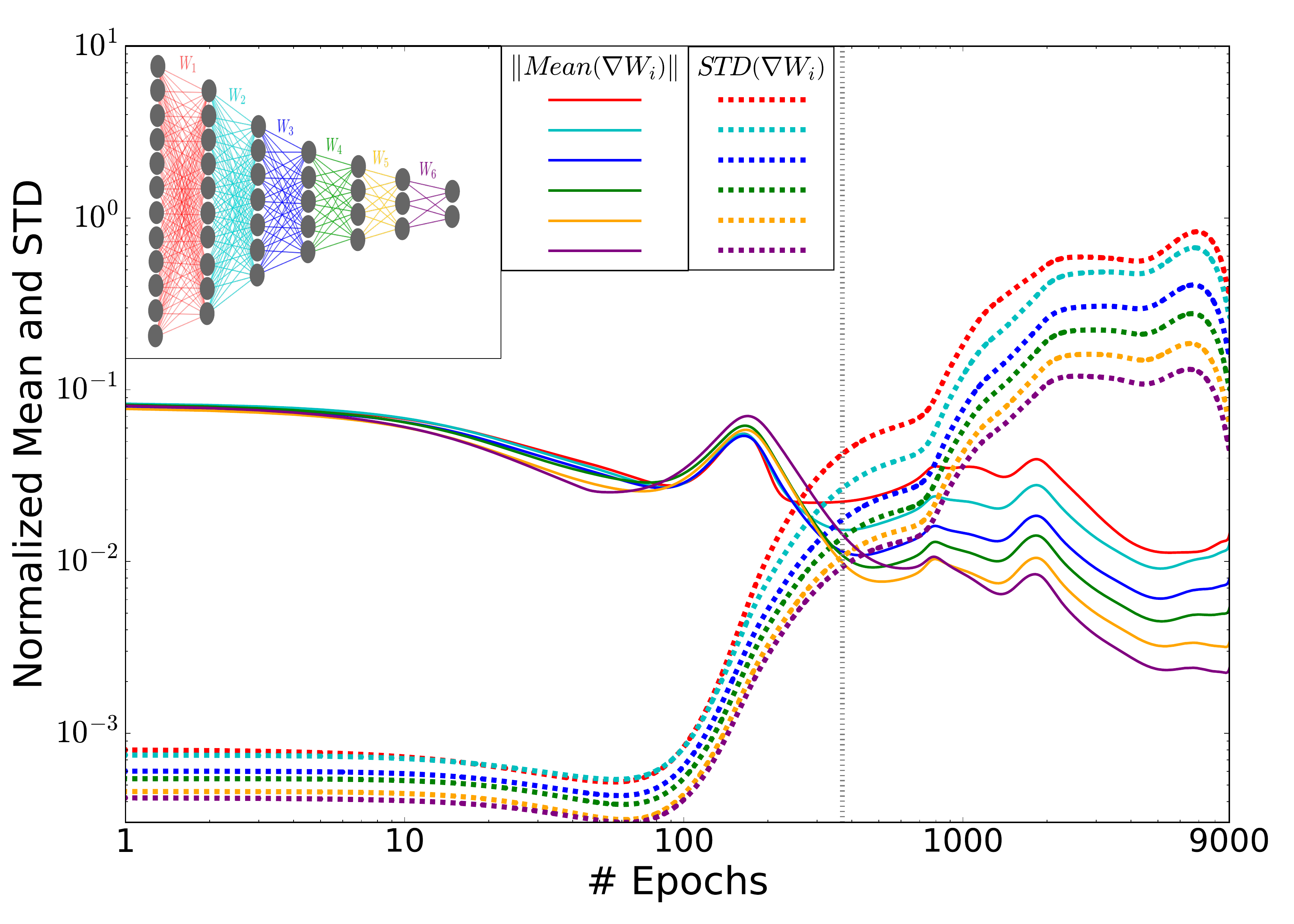}}
\caption{\textbf{The layers' Stochastic Gradients distributions during the optimization process.} The norm of the means and standard deviations of the weights gradients for each layer, as function of the number of training epochs (in log-log scale). The values are normalized by the L2 norms of the weights for each layer, which significantly increase during the optimization. The grey line ($\sim350$ epochs) marks the transition between the first phase, with large gradient means and small variance (\emph{drift}, high gradient SNR), and the second phase, with large fluctuations and small means (\emph{diffusion}, low SNR). Note that the gradients log (SNR) (the log differences between the mean and the STD lines) approach a constant for all the layers, reflecting the convergence of the network to a configuration with constant flow of relevant information through the layers!    
%The fastest drift occurs when enough information flows through the layers (around 200 epochs). 
}
\label{fig:gradients}
\vskip -0.1in
\end{figure}

A better understanding of the ERM and representation-compression phases can be derived from examination of the behavior of the stochastic gradients along the epochs. In Figure \ref{fig:gradients} we plot the normalized mean and standard deviations of the weights' stochastic gradients (in the samples batches), for every layer of our DNN (shown in the inset), as function of the SG epochs. Clearly there is a transition between two distinct phases (the vertical line). The first is a \emph{drift phase}, where the gradient means are much larger than their standard deviations, indicating small gradient stochasticity (high SNR). In the second phase, the gradient means are very small compared to their batch to batch fluctuations, and the gradients behave like Gaussian noise with very small means, for each layer (low SNR). We call this the \emph{diffusion phase}. Such a transition is expected in general, when the empirical error saturates and SGD is dominated by its fluctuations.  
We claim that these distinct SG phases (grey line in Figure \ref{fig:gradients}), correspond and explain the ERM and compression phases we observe in the \textit{information plane} (green paths marked on the layers' trajectories in Figure \ref{network_epochs}).  

This dynamic phase transition occurs in the same number of epochs as the left bent of the layers' trajectories in the \textit{information plane}. The drift phase clearly increases $I_Y$ for every layer, since it quickly reduces the empirical error. On the other hand, the diffusion phase mostly adds random noise to the weights, and they evolve like Wiener processes, under the training error or label information constraint. Such diffusion processes can be described by a Focker-Planck equation [see e.g. \citet{risken1989fokker}], whose stationary distribution maximizes the entropy of the weights distribution, under the training error constraint.  That in turn maximizes the conditional entropy, $H(X|T_i)$, or minimizes the mutual information $I(X;T_i)=H(X)-H(X|T_i)$, because the input entropy, $H(X)$, does not change.  This entropy maximization by additive noise, also known as stochastic relaxation, is constrained by the empirical error, or equivalently (for small errors) by the $I_Y$ information.  
We present a rigorous analysis of this stochastic relaxation process elsewhere, but it is already clear how the diffusion phase can lead to more compressed representations, by minimizing $I_X$  for every layer.     

However, it remains unclear why different hidden layers converge to different points in the \textit{information plane}. Figure \ref{fig:gradients} suggests that different layers have different levels of noise in the gradients during the compression phase, which can explain why they end up in different maximum entropy distributions. But as the gradient noises seem to vary and eventually decrease when the layers converge, suggesting that the convergence points are related to the \emph{critical slowing down} 
\ignore{\ravidcomment{I think we need to say what is critical slowing down, maybe -  critical slowing down is a phenomenon where  system dynamics should recover more slowly from small perturbations upon approaching the bifurcation or tipping point. Critical slowing down can be monitored by measuring the recovery rate of system variables after small perturbations but also manifests itself by an increase in its fluctuations, i.e. variance due to the longer relaxation times near the bifurcation} [\citet{wissel1984universal}, \citet{ives1995measuring}] } 
of stochastic relaxation near phase transitions on the Information Bottleneck curve. 
This intriguing hypothesis is further examined elsewhere. 

Another interesting consequence of the \emph{compression by diffusion phase} is the randomized nature of the final weights of the DNN. We found no indication for vanishing connections or norm decreases near the convergence. This
is consistent with previous works which showed that explicit forms of regularization, such as weight decay, dropout, and data augmentation, do not adequately explain the generalization error of DNNs [\citet{DBLP:journals/corr/ZhangBHRV16}]. Moreover, the correlations between the in-weights of different neurons in the same layer, which converge to essentially the same point in the plane, was very small. This indicates that there is a huge number of different networks with essentially optimal performance, and \emph{attempts to interpret single weights or even single neurons in such networks can be meaningless. }
%\todo{Insert here the norms of the weights figures? To show that they are not change? also need to explain it better}

\subsection{The computational benefit of the hidden layers}

We now turn to one of the fundamental questions about Deep Learning - what is the benefit of the hidden layers? 

\begin{figure}[t]
\begin{centering}
\includegraphics[width=\textwidth, scale = 0.5]{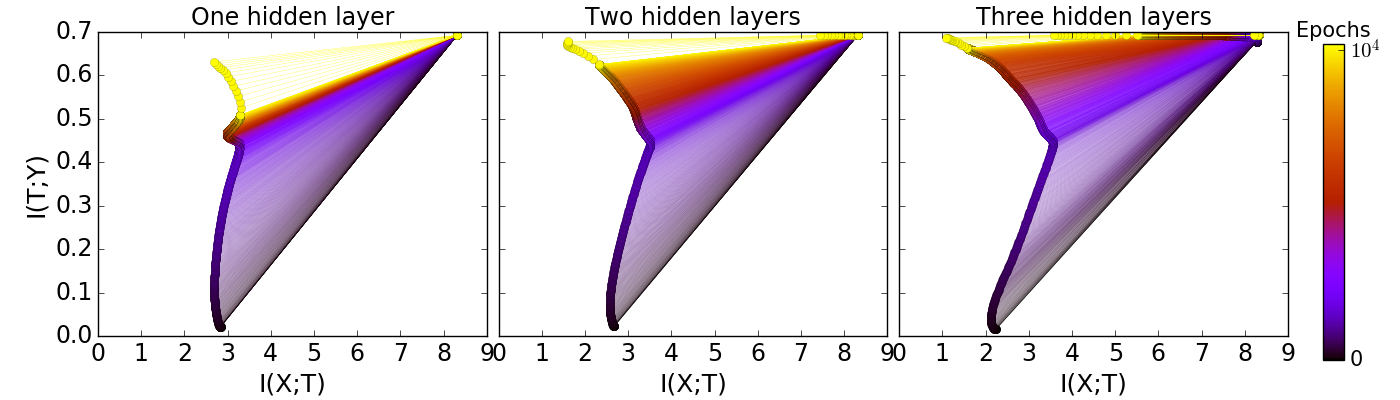}
\includegraphics[width=\textwidth, scale = 0.5]{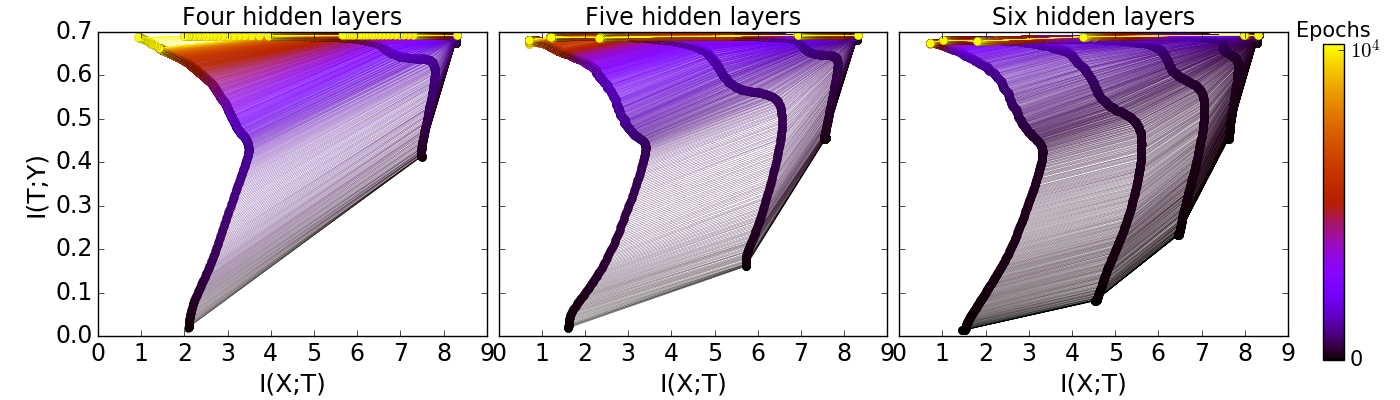}
\par\end{centering}
\caption{\textbf{The layers information paths during the SGD
optimization for different architectures.} 
Each panel is the \textit{information plane} 
for a network with a different number of hidden
layers. The width of the hidden layers start with 12, and each additional layer has 2 fewer neurons. 
The final layer with 2 neurons is shown in all panels.  
The line colors correspond to the number of training epochs. 
%The green line - $H\left(Y\right)$ is the theoretical bound for $I_{Y}$.
}
\vskip -0.1in
\label{layers_inf}
\end{figure}

To address this, we trained 6 different architectures with
1 to 6  hidden layers (with layers as in Figure \ref{fig:gradients}), trained on 80\%  
the patterns, randomly sampled from Eq.(\ref{eq:rule}). 
As before, we repeated each training 50 times with randomized initial weights and training samples. 
Figure \ref{layers_inf} shows the information plane paths for these 6 architectures during the training epochs, each averaged over the randomized networks. 

There are several important outcomes of this experiment:

1. \emph{Adding hidden layers dramatically reduces the number of training epochs for good generalization.}

To see this, compare the color of the paths at the first panels of Figure \ref{layers_inf} (with 1 and 2 hidden layers), with the colors in the last panels (with 5 and 6 hidden layers). Whereas with 1 hidden layer the network was unable to achieve good $I_Y$ values even after $10^4$ epochs, with 6 hidden layers it reached the full relevant information at the output layer within 400 epochs.
  
2. \emph{The compression phase of each layer is shorter when it starts from a previous compressed layer.}

This can be seen by comparing the time to good generalization with 4 and 5 hidden layers. The yellow at the top indicates a much slower convergence with 4 layers than with 5 or 6 layers, where they reach the end points with half the number of epochs. 

3. \emph{The compression is faster for the deeper (narrower and closer to the output) layers.}

Whereas in the drift phase the lower layers move first (due to DPI), in the diffusion phase the top layers compress first and "pull" the lower layers after them. Adding more layers seems to add intermediate representations which accelerates the compression.   
%\mycomment{Ravid? check this, ravid:True}
%\ignore{
4. \emph{Even wide hidden layers eventually compress in the diffusion phase. Adding extra width does not help.}

It is clear from panels 4, 5 and 6 that the first hidden layer (of width 12) remains in the upper right corner, without lose of information on either $X$ or $Y$. Our simulations suggest that such 1-1 transformations do not help learning, since they do not generate compressed representations. Others have suggested that such layers can still improve the Signal-to-noise ratio (SNR) of the patterns in some learning models [\citet{Kadmon2016OptimalAI}]. In our simulations all the hidden layers eventually compress the inputs, given enough SGD epochs.  
%}

\subsection{The computational benefits of layered diffusion}
\label{com.benefit}
Diffusion processes are governed by the diffusion equation, or by the Focker-Planck equation if there is also a drift or a constraining potential.  In simple diffusion, the initial distribution evolves through convolution with a Gaussian kernel, whose width grows like $\sqrt{Dt}$ with time, in every dimension ($D$ - a diffusion constant). Such convolutions lead to  an entropy increase which grows like $\Delta H \propto  \log (Dt)$. Thus the entropy growth is logarithmic in the number of time steps, or the number of steps is exponential in the entropy growth.  If there is a potential, or empirical error constraint, this process converges asymptotically to the maximum entropy distribution, which is exponential in the constrained potential or training error. This exponential distribution meets the IB equations Eq. (\ref{eqn:IB}), as we saw in section \ref{IB.optimal}.

When applying this to the diffusion phase of the SGD optimization in DNN, one can expect a compression $\Delta I_X$ by diffusion to be of order $\exp(\Delta I_X/D)$ time steps, or optimization epochs.  Assume now that with $K$ hidden layers, each layer only needs to compress by diffusion from the previous (compressed) layer, by  $\Delta I^{k}_X$. One can see that the total compression, or entropy increase, approximately breaks down into $K$ smaller steps, $\Delta I_X \approx \sum_k \Delta  I^{k}_X$. As
\begin{equation}
\exp (\sum_k \Delta  I^{k}_X) \gg \sum_k \exp(\Delta  I^{k}_X) ~,
\end{equation}
 there is an exponential (in the number of layers $K$, if the $\Delta  I^{k}_X$ are similar) decrease in epochs with $K$ hidden layers. Note that if we count operations, they only grow linearly with the number of layers, so this exponential boost in the number of epochs can still be very significant. This remains true as long as the number of epochs is super-linear in the compressed entropy.

\subsection{Convergence to the layers to the Information Bottleneck bound}
\label{IB.optimal}

In order to quantify the IB optimality of the layers we tested whether the converged layers satisfied the encoder-decoder relations of Eq. (\ref{eqn:IB}), for some value of the Lagrange multiplier $\beta$.
For each converged layer we used the encoder and decoder distributions based on
the layer neurons' quantized values, $p_{i}\left(t|x\right)$ and $p_{i}\left(y|t\right)$
with which we calculated the information values $\left(I_{X}^{i},I_{Y}^{i}\right)$. 

To test the IB optimality of the layers encoder-decoder we calculated the optimal IB encoder, 
$p_{i,\beta}^{IB}\left(t|x\right)$
using the $i^{th}$ layer decoder, $p_{i}\left(y|t\right)$, through Eq.(\ref{eqn:IB}). 
This can be done for any value of $\beta$, with the known $P(X,Y)$. 

We then found the optimal $\beta_i$ for each layer, by minimizing the averaged KL divergence between the IB and the layer's encoders,
\begin{equation}
\beta_{i}^{\star}=\arg \min_{\beta} \mathbb{E}_{x}  { D_{KL}\left[p_{i}\left(t|x\right)||p_{\beta}^{IB}\left(t|x\right)\right]} ~.
\end{equation}
\begin{figure}[ht]
\centering
\includegraphics[width=0.95\textwidth]{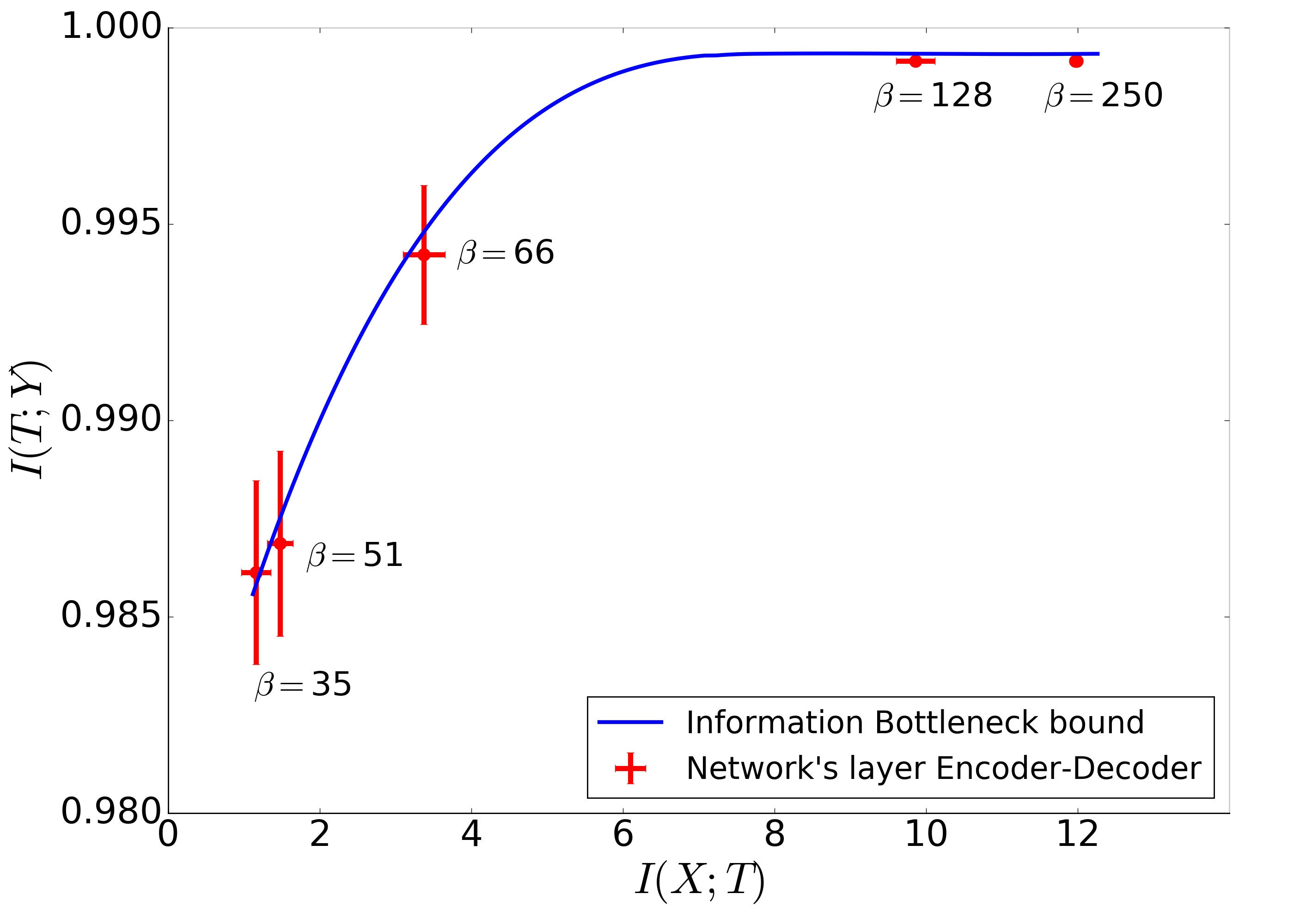}
\caption{\textbf{\label{IB}
The DNN layers converge to fixed-points of the IB equations}. 
The error bars represent standard error measures with N=50. In each line
there are 5 points for the different layers. For each point, $\beta$
is the optimal value that was found for the corresponding layer.}
\vskip -0.2in
\end{figure}

In Figure \ref{IB} we plot the \textit{information plane} with the layers' information values $\left(I_{X}^{i},I_{Y}^{i}\right)$ and the IB information curve (blue line). The 5 empirical layers (trained with SGD) lie remarkably close to the theoretical IB limit, where the slope of the curve, $\beta^{-1}$, matches their estimated optimal $\beta_{i}^{\star}$.   
%\mycomment{Note that in this procedure we replaced the IB $\hat{t}$ representations by the DNN $i^{th}$ layer $t_i$, and only then could calculate this KL by averaging over the possible values of the layer $t$. This requires an explanation. It can be done  because in the IB equations the representations $T$ or $\hat{X}$ can be anything, up to information preservibg maps. This is the main power of the IB! We need to make it more explicit in the final or extended version. It is important since the layers and the IB representations are very different mathematical concepts.  }

Hence, the DNN layers' encoder-decoder distributions satisfy the IB self-consistent
equations within our numerical precision, with decreasing  $\beta$ as we move to deeper layers.  
The error bars are calculated over the 50 randomized networks. 
As predicted by the IB equations, near the information curve $\Delta I_Y \sim \beta^{-1} \Delta I_X$. 
How exactly the DNN neurons capture the optimal IB representations is another interesting issue to be discussed elsewhere, but there are clearly many different layers that correspond to the same IB representation.

\subsection{Evolution of the layers with training sample size}

\begin{figure}[h]
\begin{centering}
\includegraphics[scale = 0.65]{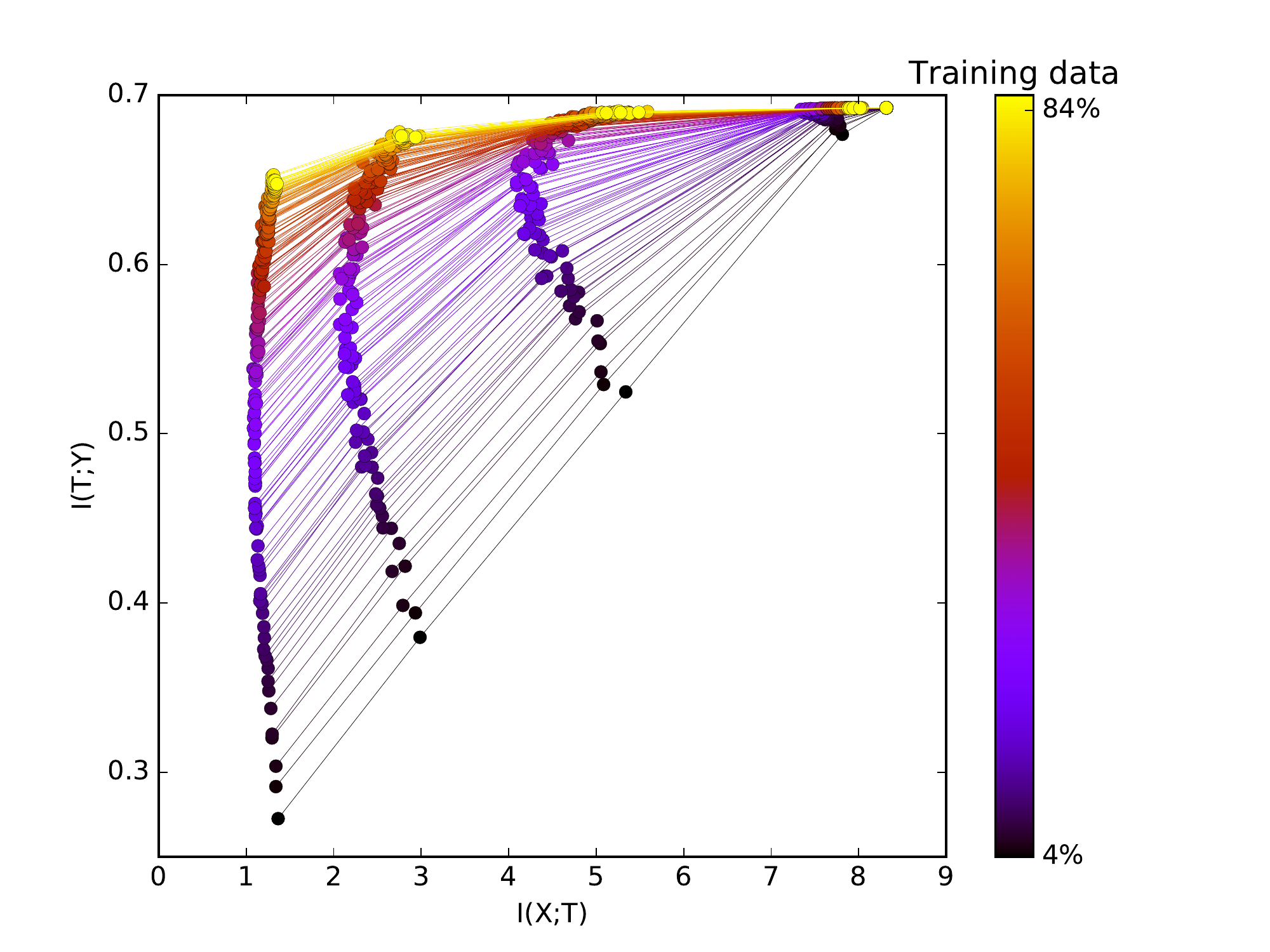}
\par\end{centering}
\caption{\label{samples_layers}\textbf{The effect of the training data size
on the layers in the \textit{information plane}.} Each line (color) represents a converged network with a different training sample size. Along each line there are 6 points for the different layers, each averaged over 50 random training samples and randomized initial weights.}
\end{figure}

Another fundamental issue in machine learning, which we only deal with briefly in this paper, is the dependence on the training sample size.
\ignore{\ravidcomment{I add some reference of someone that explores the accuracy as function the data sizes in datasets of images, do you think we need to keep it? }} [\citet{cho2015much}].
It is useful to visualize the converged locations of the hidden layers for different training data sizes
in the \textit{information plane} (Figure \ref{samples_layers}).

\vskip -0.1in 

We trained networks with 6 hidden layers as before, but with different sample sizes, ranging from 3\% to 85\% of the patterns.  
As expected, with increasing training size the layers' true label information (generalization) $I_Y$ is pushed up and 
gets closer to the theoretical IB bound for the rule distribution. 

Despite the randomizations, the converged layers for different training sizes
lie on a smooth line for each layer, with remarkable regularity. We claim that the 
layers converge to specific points on the finite sample information curves, which can be calculated using the IB self-consistent equations (Eq. (\ref{eqn:IB})), with the decoder replaced by the empirical distribution. This finite sample IB bound also explains the bounding shape on the left of Figure \ref{network_epochs}.   
Since the IB information curves are convex for any distribution, even with very small samples the layers converge to a convex curve in the plane. 
%More details will be given elsewhere. 

The effect of the training size on the layers is different for $I_{Y}$ and $I_{X}$. 
In the lower layers, the training size hardly changes the information at all, since even random weights keep most of the mutual information on both $X$ and $Y$. However, for the deeper layers the
network learns to preserve more of the information on $Y$ and better compress the irrelevant information in $X$. 
With larger training samples more details on $X$ become relevant for $Y$ and we there is a shift to higher $I_X$ in the middle layers.

\section{Discussion}
\label{Sec:discussion}
Our numerical experiments  were motivated by the Information Bottleneck framework. 
We demonstrated that the visualization of the layers in the \textit{information plane} reveals many - so far unknown - details about the inner working of Deep Learning and Deep Neural Networks. They revealed the distinct phases of the SGD optimization, drift and diffusion, which explain the ERM and the representation compression trajectories of the layers. The stochasticity of SGD methods is usually motivated as a way of escaping local minima of the training error. In this paper we give it a new, perhaps much more important role: it generates highly efficient internal representations through \emph{compression by diffusion}. This is consistent with other recent suggestions on the role of noise in Deep Learning [\citet{2016arXiv161101353A}, \citet{Kadmon2016OptimalAI}, \citet{DBLP:journals/corr/BalduzziFLLMM17}].
%They suggest a new theory for the huge computational benefit of the hidden layers, which we outline below. 

\begin{figure}[ht]
\begin{center}
\begin{minipage}[c]{0.5\linewidth}
\includegraphics[width=\textwidth,height=5.8cm]{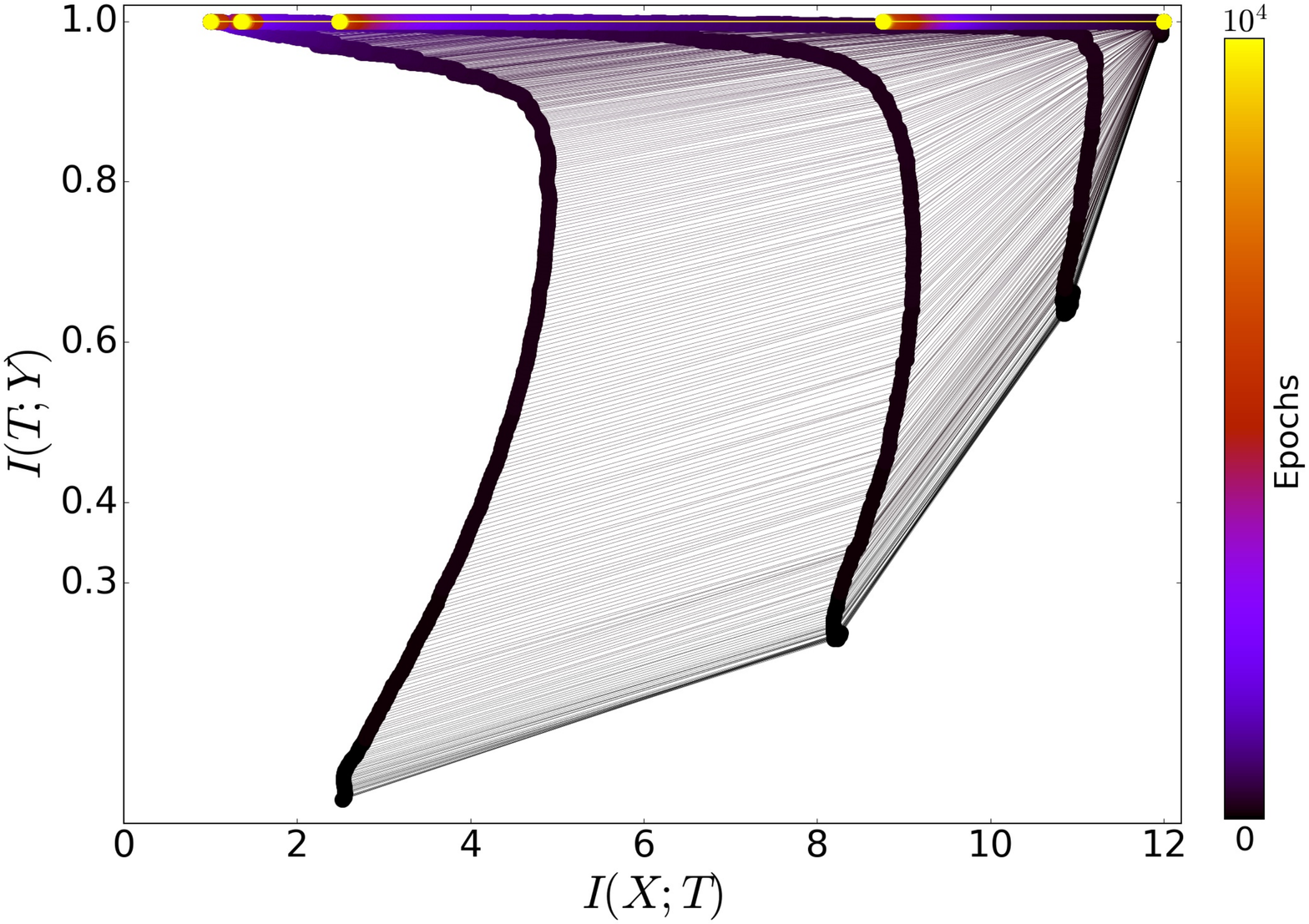}
\end{minipage}%\haspace{fill}
\begin{minipage}[c]{0.5\linewidth}
\includegraphics[width=\textwidth,height=5.8cm, scale = 0.5]{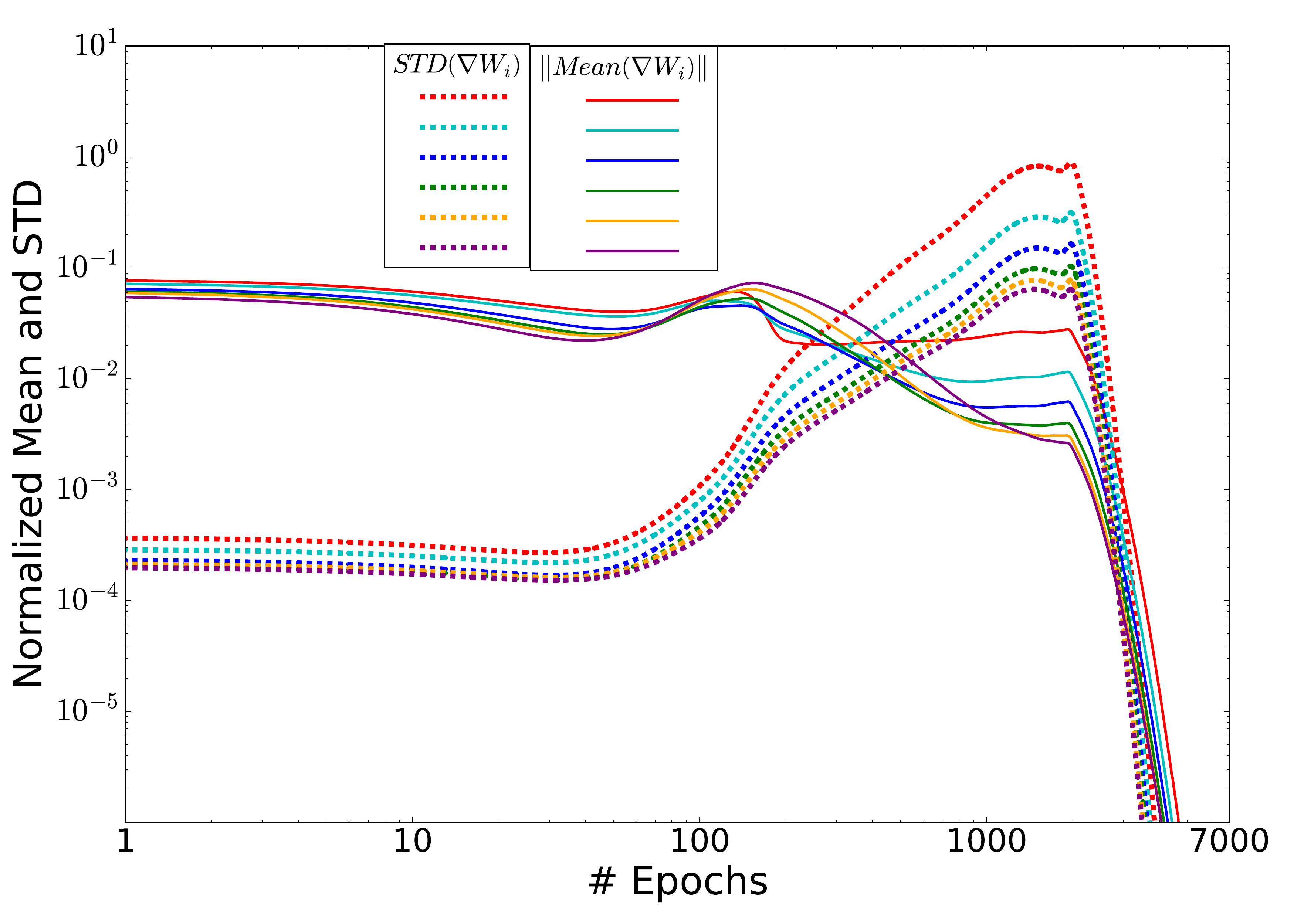}
\end{minipage}

\end{center}
\caption{\textbf{The layers information plane paths (left) and stochastic gradients means and standard deviations for a non-symmetric committee machine rule.} Clearly seen are the two phases of the optimization process as in the symmetric rule. One can also see the equilibration of the gradient SNR for the different layers. While the compression phase is faster in this case, the overall training dynamics is very similar.}
\label{committee-machine}
\vskip -0.1in
\end{figure}

Some critical obvious questions should be discussed. 
\begin{enumerate}
\item Are our findings general enough? Do they occur with other rules and network architectures? 
\item Can we expect a similar compression by noise phase in other, not DNN, learning models? 
\item Do they scale up to larger networks and "real world" problems?
\item What are the practical or algorithmic implications of our analysis and findings?
\end{enumerate}
To answer the first question, we repeated our information plane and stochastic gradient analysis with an entirely different non-symmetric rule and architectures - the well studied committee machine [e.g. \citet{Haykin:1998:NNC:521706}]. As can be seen in figure \ref{committee-machine} the Information Plane paths and the SGD phases are very similar, and exhibit essentially the same diffusion and compression phase, with similar equilibration of the relevant information channels.

The second question is interesting. It is well known that in a single layer networks (perceptron) the generalization error is the $arc-cosine$ of the scalar product of the "teacher" and "student" weights. In that case adding noise to the weights can only decrease the generalization error (on average). So our generalization through noise mechanism depends on the multi-layer structure of the network, but may occur in other "deep" models, such as Bayesian networks, or random forests.     

For the third question, we examined the the SG statistics for a standard large problem (MNIST classification) and found the transition to the SGD diffusion phase. Similar effects of the noise in the gradients have been recently reported also in \citet{2016arXiv161101353A} and \citet{DBLP:journals/corr/BalduzziFLLMM17}. Our analysis suggests that this is enough for the occurance of the compression phase. Direct estimation of the Information Plane paths for large problems require more sophisticated mutual information estimators, but this can be done. 

The forth question is certainly the most important for the applications of DL, and we are currently working on new learning algorithms that utilize the claimed IB optimality of the layers. We argue that SGD seems an overkill during the diffusion phase, which consumes most of the training epochs, and that much simpler optimization algorithms, such as Monte-Carlo relaxations [\citet{geman1988stochastic}], can be more efficient.  

But the IB framework may provide even more. If the layers actually converge to the IB theoretical bounds, there is an analytic connection between the encoder and decoder distributions for each layer, which can be exploited during training. Combining the IB iterations with stochastic relaxation methods may significantly boost DNN training.

%\subsection{The network layers and the Information Bottleneck critical points}

To conclude, it seems fair to say, based on our experiments and analysis, that Deep Learning with DNN 
are in essence learning algorithms that effectively find efficient representations that are approximate minimal sufficient statistics in the IB sense.  

If our findings hold for general networks and tasks, the compression phase of the SGD and the convergence of the layers to the IB bound can explain the phenomenal success of Deep Learning.

%{\noindent \em Remainder omitted in this sample. See http://www.jmlr.org/papers/ for full paper.}

% Acknowledgements should go at the end, before appendices and references

\subsection*{Acknowledgements} 
This study was supported by the Israeli Science Foundation center of excellence, the
Intel Collaborative Research Institute for Computational Intelligence (ICRI-CI),
and the Gatsby Charitable Foundation. 


\appendix
%\section*{Appendix A.}
\label{app:theorem}

% Note: in this sample, the section number is hard-coded in. Following
% proper LaTeX conventions, it should properly be coded as a reference:

\vskip 0.5in
%\newpage

\bibliography{deepIB}

\begin{thebibliography}{22}
\providecommand{\natexlab}[1]{#1}
\providecommand{\url}[1]{\texttt{#1}}
\expandafter\ifx\csname urlstyle\endcsname\relax
  \providecommand{\doi}[1]{doi: #1}\else
  \providecommand{\doi}{doi: \begingroup \urlstyle{rm}\Url}\fi

\bibitem[{Achille} and {Soatto}(2016)]{2016arXiv161101353A}
A.~{Achille} and S.~{Soatto}.
\newblock {Information Dropout: Learning Optimal Representations Through Noisy
  Computation}.
\newblock \emph{ArXiv e-prints}, November 2016.

\bibitem[Alain and Bengio(2016)]{probes2016}
Guillaume Alain and Yoshua Bengio.
\newblock Understanding intermediate layers using linear classifier probes,
  2016.

\bibitem[Balduzzi et~al.(2017)Balduzzi, Frean, Leary, Lewis, Ma, and
  McWilliams]{DBLP:journals/corr/BalduzziFLLMM17}
David Balduzzi, Marcus Frean, Lennox Leary, J.~P. Lewis, Kurt~Wan{-}Duo Ma, and
  Brian McWilliams.
\newblock The shattered gradients problem: If resnets are the answer, then what
  is the question?
\newblock \emph{CoRR}, abs/1702.08591, 2017.
\newblock URL \url{http://arxiv.org/abs/1702.08591}.

\bibitem[Cho et~al.(2015)Cho, Lee, Shin, Choy, and Do]{cho2015much}
Junghwan Cho, Kyewook Lee, Ellie Shin, Garry Choy, and Synho Do.
\newblock How much data is needed to train a medical image deep learning system
  to achieve necessary high accuracy?
\newblock \emph{arXiv preprint arXiv:1511.06348}, 2015.

\bibitem[Cover and Thomas(2006)]{Cover:2006}
Thomas~M. Cover and Joy~A Thomas.
\newblock \emph{Elements of Information Theory (Wiley Series in
  Telecommunications and Signal Processing)}.
\newblock Wiley-Interscience, 2006.

\bibitem[Geman and Geman(1988)]{geman1988stochastic}
Stuart Geman and Donald Geman.
\newblock Stochastic relaxation, gibbs distributions, and the bayesian
  restoration of images, neurocomputing: foundations of research, 1988.

\bibitem[Graves et~al.(2013)Graves, Mohamed, and
  Hinton]{DBLP:journals/corr/abs-1303-5778}
Alex Graves, Abdel-rahman Mohamed, and Geoffrey Hinton.
\newblock Speech recognition with deep recurrent neural networks.
\newblock In \emph{Acoustics, speech and signal processing (icassp), 2013 ieee
  international conference on}, pages 6645--6649. IEEE, 2013.

\bibitem[Haykin(1998)]{Haykin:1998:NNC:521706}
Simon Haykin.
\newblock \emph{Neural Networks: A Comprehensive Foundation}.
\newblock Prentice Hall PTR, Upper Saddle River, NJ, USA, 2nd edition, 1998.
\newblock ISBN 0132733501.

\bibitem[He et~al.(2015)He, Zhang, Ren, and Sun]{DBLP:journals/corr/HeZRS15}
Kaiming He, Xiangyu Zhang, Shaoqing Ren, and Jian Sun.
\newblock Deep residual learning for image recognition.
\newblock \emph{CoRR}, abs/1512.03385, 2015.

\bibitem[Hinton et~al.(2012)Hinton, Srivastava, Krizhevsky, Sutskever, and
  Salakhutdinov]{DBLP:journals/corr/abs-1207-0580}
Geoffrey~E Hinton, Nitish Srivastava, Alex Krizhevsky, Ilya Sutskever, and
  Ruslan~R Salakhutdinov.
\newblock Improving neural networks by preventing co-adaptation of feature
  detectors.
\newblock \emph{arXiv preprint arXiv:1207.0580}, 2012.

\bibitem[Kadmon and Sompolinsky(2016)]{Kadmon2016OptimalAI}
Jonathan Kadmon and Haim Sompolinsky.
\newblock Optimal architectures in a solvable model of deep networks.
\newblock In \emph{NIPS}, 2016.

\bibitem[Kazhdan et~al.(2003)Kazhdan, Funkhouser, and
  Rusinkiewicz]{Kazhdan2003}
Michael Kazhdan, Thomas Funkhouser, and Szymon Rusinkiewicz.
\newblock Rotation invariant spherical harmonic representation of 3d shape
  descriptors.
\newblock \emph{Eurographics Symposium on Geometry Processing}, 2003.

\bibitem[Kraskov et~al.(2004)Kraskov, St\"ogbauer, and
  Grassberger]{PhysRevE.69.066138}
Alexander Kraskov, Harald St\"ogbauer, and Peter Grassberger.
\newblock Estimating mutual information.
\newblock \emph{Phys. Rev. E}, 69:\penalty0 066138, Jun 2004.
\newblock \doi{10.1103/PhysRevE.69.066138}.

\bibitem[Larochelle et~al.(2009)Larochelle, Bengio, Louradour, and
  Lamblin]{Larochelle:2009:EST:1577069.1577070}
Hugo Larochelle, Yoshua Bengio, J{\'e}r\^{o}me Louradour, and Pascal Lamblin.
\newblock Exploring strategies for training deep neural networks.
\newblock \emph{J. Mach. Learn. Res.}, 10:\penalty0 1--40, June 2009.
\newblock ISSN 1532-4435.

\bibitem[LeCun et~al.(2015)LeCun, Bengio, and Hinton]{natureDeepLeraning}
Yann LeCun, Yoshua Bengio, and Geoffrey Hinton.
\newblock Deep learning.
\newblock \emph{Nature}, 2015.

\bibitem[Moshkovich and Tishby(2017)]{MoshkovichTishby17}
Michal Moshkovich and Naftali Tishby.
\newblock Mixing complexity and its applications to neural networks.
\newblock 2017.
\newblock URL \url{https://arxiv.org/abs/1703.00729}.

\bibitem[Paninski(2003)]{Paninski:2003:EEM:795523.795524}
Liam Paninski.
\newblock Estimation of entropy and mutual information.
\newblock \emph{Neural Comput.}, 15\penalty0 (6):\penalty0 1191--1253, June
  2003.
\newblock ISSN 0899-7667.
\newblock \doi{10.1162/089976603321780272}.

\bibitem[Risken(1989)]{risken1989fokker}
H.~Risken.
\newblock \emph{The Fokker-Planck Equation: Methods of Solution and
  Applications}.
\newblock Number isbn{9780387504988}, lccn={89004059} in Springer series in
  synergetics. Springer-Verlag, 1989.

\bibitem[Tishby and Zaslavsky(2015)]{DBLP:journals/corr/TishbyZ15}
Naftali Tishby and Noga Zaslavsky.
\newblock Deep learning and the information bottleneck principle.
\newblock In \emph{Information Theory Workshop (ITW), 2015 IEEE}, pages 1--5.
  IEEE, 2015.

\bibitem[Tishby et~al.(1999)Tishby, Pereira, and
  Bialek]{DBLP:journals/corr/Tishby1999}
Naftali Tishby, Fernando~C. Pereira, and William Bialek.
\newblock The information bottleneck method.
\newblock \emph{In Proceedings of the 37-th Annual Allerton Conference on
  Communication, Control and Computing}, 1999.

\bibitem[Zhang et~al.(2016)Zhang, Bengio, Hardt, Recht, and
  Vinyals]{DBLP:journals/corr/ZhangBHRV16}
Chiyuan Zhang, Samy Bengio, Moritz Hardt, Benjamin Recht, and Oriol Vinyals.
\newblock Understanding deep learning requires rethinking generalization.
\newblock \emph{arXiv preprint arXiv:1611.03530}, 2016.

\bibitem[Zhang and LeCun(2015)]{DBLP:journals/corr/ZhangL15}
Xiang Zhang and Yann LeCun.
\newblock Text understanding from scratch.
\newblock \emph{arXiv preprint arXiv:1502.01710}, 2015.

\end{thebibliography}

\end{document}